\pdfoutput=1

\documentclass[11pt]{article}

\usepackage{acl}

\usepackage{emoji}

\usepackage{times}
\usepackage{latexsym}

\usepackage[T1]{fontenc}

\usepackage[utf8]{inputenc}

\usepackage{microtype}

\usepackage{amsmath}
\usepackage{amsfonts}       %
\usepackage{bbm}
\usepackage{enumerate}
\usepackage{enumitem,kantlipsum}
\usepackage{inconsolata}
\usepackage{cleveref}
\usepackage{mathtools}
\usepackage{xspace}
\newcommand{\XXTASKS}{28\xspace}

\newcommand{\bert}{\textsc{BERT}\xspace}
\newcommand{\tfive}{\textsc{T5}\xspace}

\newcommand{\fasttext}{fastText\xspace}

\newcommand{\property}{\ensuremath{\pi}}
\newcommand{\vproperty}{\ensuremath{\vpi}}

\newcommand{\probe}{f}
\newcommand{\vprobe}{\boldsymbol{f}}
\newcommand{\token}{\tau}
\newcommand{\vtoken}{\ensuremath{\vtau}}

\newcommand{\wordrep}{\ensuremath{h}}

\newcommand{\vwordrep}{\ensuremath{\boldsymbol h}}

\newcommand{\defn}[1]{\textbf{#1}}

\newcommand{\evidence}{evidence\xspace}

\DeclareMathOperator*{\argmax}{argmax}

\newcommand{\R}{\mathbb{R}}

\def\sP{{\mathcal{P}}}

\def\sR{{\mathcal{R}}}

\def\sV{{\mathcal{V}}}

\def\vtau{{\boldsymbol{\tau}\xspace}}
\def\vpi{{\boldsymbol{\pi}\xspace}}

\newcommand{\given}{\mid}
\newcommand{\ra}{\mbox{$\rightarrow$}}

\newcommand{\commit}{CB\xspace}
\newcommand{\rte}{RTE\xspace}
\newcommand{\boolq}{BoolQ\xspace}

\usepackage[T5,T1]{fontenc}
\usepackage{combelow}
\usepackage[utf8]{inputenc}

\DeclareTextSymbolDefault{\ohorn}{T5}
\DeclareTextSymbolDefault{\uhorn}{T5}

\crefname{section}{\S}{\S\S}
\Crefname{section}{\S}{\S\S}
\crefname{table}{Tab.}{}
\crefname{figure}{Fig.}{}
\crefname{algorithm}{Algorithm}{}
\crefname{equation}{eq.}{}
\crefname{appendix}{App.}{}
\crefname{thm}{Theorem}{}
\crefname{prop}{Proposition}{}
\crefname{cor}{Corollary}{}
\crefname{observation}{Observation}{}
\crefname{assumption}{Assumption}{}
\crefformat{section}{\S#2#1#3}

\usepackage{bbm}
\usepackage{todonotes}
\makeatletter
\newcommand*\iftodonotes{\if@todonotes@disabled\expandafter\@secondoftwo\else\expandafter\@firstoftwo\fi}  %
\makeatother

\definecolor{dandelion}{HTML}{FFD464}

\usepackage{caption,subcaption}
\usepackage{booktabs}
\usepackage{tikz}
\usetikzlibrary{positioning}
\usetikzlibrary{calc}
\definecolor{lightgray}{gray}{0.85}
\definecolor{lightlightgray}{gray}{0.9}
\definecolor{C1}{HTML}{1F77B4}
\definecolor{C2}{HTML}{FF7F0E}
\definecolor{C3}{HTML}{2CA02C}
\definecolor{C4}{HTML}{D62728}
\definecolor{C5}{HTML}{9467BD}

\everypar={\looseness=-1}

\title{Probing as Quantifying Inductive Bias}

\usepackage{tipa}

\newcommand{\ethz}{ \emoji[twitter-emoji]{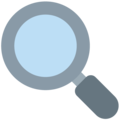}}
\newcommand{\mitinst}{\emoji[twitter-emoji]{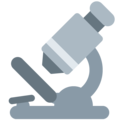}}

\newcommand{\customfootnotetext}[2]{{%
  \renewcommand{\thefootnote}{#1}%
  \footnotetext[0]{#2}}}%
  
\author{
Alexander Immer$^{*, \ethz}$~\;~Lucas Torroba Hennigen$^{*, \mitinst}$ ~\;~\textbf{Vincent Fortuin}$^{\ethz}$~\;~\textbf{Ryan Cotterell}$^{\ethz}$
\\
$^{\ethz}$ETH Z{\"u}rich \quad $^{\mitinst}$MIT \\
\href{mailto:alexander.immer@inf.ethz.ch}{\texttt{alexander.immer@inf.ethz.ch}}~\;~ \href{mailto:lucastor@mit.edu}{\texttt{lucastor@mit.edu}} 
\\
\href{mailto:fortuin@inf.ethz.ch}{\texttt{fortuin@inf.ethz.ch}}~\;~ \href{mailto:ryan.cotterell@inf.ethz.ch}{\texttt{ryan.cotterell@inf.ethz.ch}}
}

\date{}

\begin{document}
\maketitle

\customfootnotetext{*}{Equal contribution.}

\begin{abstract}

Pre-trained contextual representations have led to dramatic performance improvements on a range of downstream tasks.
Such performance improvements have motivated researchers to quantify and understand the linguistic information encoded in these representations.
In general, researchers quantify the amount of linguistic information through probing, an endeavor which consists of training a supervised model to predict a linguistic property directly from the contextual representations. 
Unfortunately, this definition of probing has been subject to extensive criticism in the literature, and has been observed to lead to paradoxical and counter-intuitive results.
In the theoretical portion of this paper, we take the position that the goal of probing ought to be measuring the amount of inductive bias that the representations encode on
a specific task.
We further describe a Bayesian framework that operationalizes this goal and allows us to quantify the representations' inductive bias.
In the empirical portion of the paper, we apply our framework to a variety of NLP tasks.
Our results suggest that our proposed framework alleviates many previous problems found in probing.
Moreover, we are able to offer concrete evidence that---for some tasks---fastText can offer a better inductive bias than BERT.\footnote{Our code is available at \url{https://github.com/rycolab/evidence-probing}.}

\end{abstract}

\section{Introduction}
\label{sec:intro}

Improved pre-trained representations have led to new performance heights on NLP applications.
This has prompted researchers to analyze these representations in an attempt to determine which linguistic properties they encode.
\defn{Probing} is the primary method to perform such a quantification; typically, probing consists of training a supervised model, called a \defn{probe}, to predict a linguistic property directly from the representations.
It has been argued that the existence of a high-performing probe suggests that the representation encodes the property of interest~\citep{alain2016understanding, belinkov2018survey}.
However, despite the apparent simplicity of probing and its wide-spread use, the community has yet to find consensus on several important problems about the endeavor.
We enumerate several problems with the supervised probing framework in the following paragraphs.

\paragraph{Problem I (Representation Selection).} 
Counterintuitively, probing may fail to capture observed differences between representations. 
For instance, in some supervised probing studies, researchers have shown that random representations are equally good or better than trained ones~\citep{zhang2018auxiliary, pimentel2020pareto}.
This is certainly a nonsensical result; random representations, by construction, do not encode any linguistic property.

\paragraph{Problem II (Probe Selection).} 
There is an ongoing debate on the choice of probes: initially, linear probes were proposed to test the linear separability of learned representations~\citep{montavon2011kernel, alain2016understanding, liu2019ling}.
However, more recently, neural networks have been applied with the explicit goal of extracting as much information as possible from the representations~\citep{adi2016probing, conneau2018sentling, pimentel2020infoprobe, pimentel2021bayesian}.
Not surprisingly, it has been found that more complex probing tasks often require more complex probes~\citep{belinkov2018survey}.
To reduce the risk of overfitting, recent methods aim at trading off probing performance with the probe's complexity~\citep{hewitt2019controlprobe,pimentel2020pareto,voita2020mdlprobing}.\looseness=-1

\paragraph{Problem III (Task Selection).}
The relationship between probing tasks and NLP tasks remains unclear.
This lack of clarity manifests itself in several ways.
Firstly, while some argue that probing should focus on simple tasks~\citep{conneau2018sentling}, others argue that probing should focus on complex tasks to be informative~\citep{pimentel2020pareto}.
Thus, it is unclear where to place the boundary between probing and regular NLP tasks and whether there should even be a distinction between the two types of tasks at all.
Secondly, how researchers should interpret experimental probing results is still up for debate.
For instance, knowing that \bert excels at text generation, is it really surprising that we can predict the tense of a word from a \bert representation?
Indeed, the NLP community is still in search of how probing can be of service to downstream tasks.

This paper proposes a new framework for supervised probing that seeks to address the problems described above.
We propose to compare representations in terms of the \defn{inductive bias} they provide for a particular task.
This may seem counterintuitive, since classical machine learning often refers to the inductive biases of models alone, and not of representations; however, we propose to instead think of models as representation--probe pairs.
Such a paired model takes raw text as input, converts it into a representation, e.g., using \bert~\citep{BERT},
and predicts a property of interest using a probe.
We formalize the notion of the inductive bias of a paired model using the Bayesian model \defn{\evidence}.
The \evidence naturally trades off performance and complexity~\citep{rasmussen2001occam, mackay2003information, bishop2006pattern},
therefore, it is well-suited to quantify the amount of inductive bias that a representation--probe pair provides for a particular task.

Indeed, we argue that, by quantifying inductive biases using the evidence, we can solve the problems listed above.
The \evidence inherently penalizes random representations, addressing \textbf{Problem I}, and allows us to automatically select probes that have the right complexity for the given task and representation, addressing \textbf{Problem II}.
Importantly, automatically controlling probe complexity leads to an apples-to-apples comparison among representations, since every representation has access to the probe best suited for it.
For example, we now have a fair basis for comparison between acontextual \fasttext representations and contextual \bert representations.
Finally, \evidence-based probing unifies probing and task-driven NLP~(\textbf{Problem III}):
the goal of the experimenter should be to identify the representation--probe pair with the best inductive bias for a particular problem so there is no difference in how the framework handles probing tasks and regular NLP tasks.\looseness=-1

To validate our framework, we apply it to \XXTASKS tasks, many of which have been used for probing before.
Our results suggest that our framework provides a practical solution to \textbf{Problem I} and \textbf{Problem II}.
With respect to \textbf{Problem I}, we never find that random representations encode more inductive bias for a task than pre-trained representations.
With respect to \textbf{Problem II}, we find that the optimal choice of probe depends on the task and representation in question, e.g., when relying on random representations, a linear probe suffices (since the added complexity of a neural probe cannot possibly help); however, with \bert representations, sometimes it is better to use a non-linear probe.
This suggests that our method automatically gets around the probe selection problem.
Moreover, our results also suggest that \fasttext can provide a better inductive bias than \bert for some morphosyntactic probing tasks.

\begin{figure*}[ht]
    \centering
    \begin{tikzpicture}
    \node (left) {\includegraphics[width=0.22\textwidth]{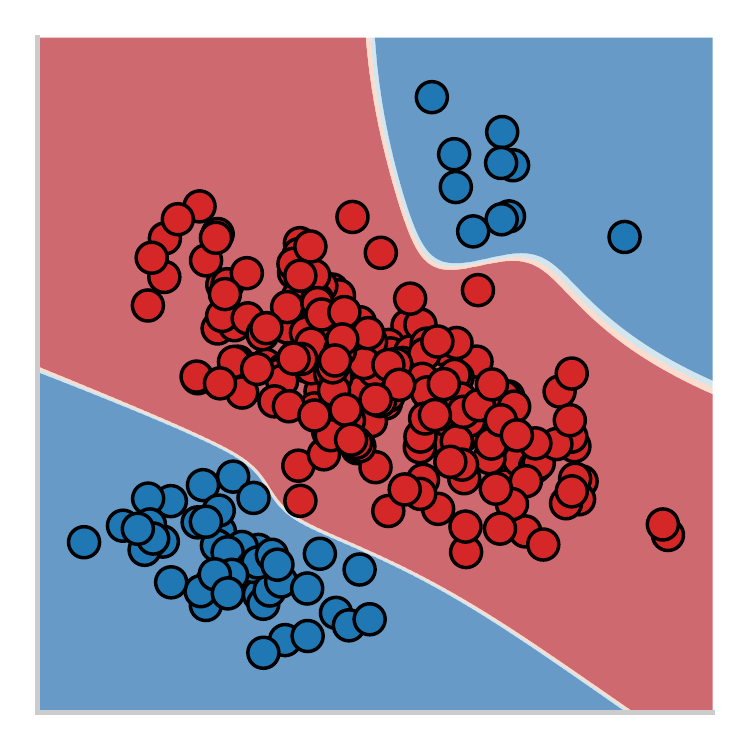}};
    \node[right=.1cm of left] (midleft){\includegraphics[width=0.22\textwidth]{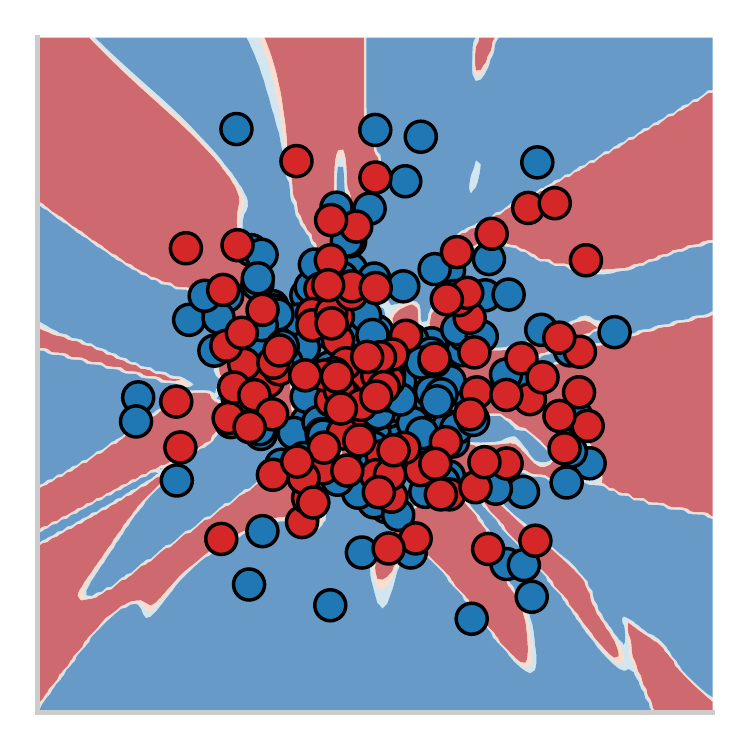}};
    \node[right=.2cm of midleft] (midright){\includegraphics[width=0.22\textwidth]{figures/mlp_bert.pdf}};
    \node[right=.1cm of midright] (right) {\includegraphics[width=0.22\textwidth]{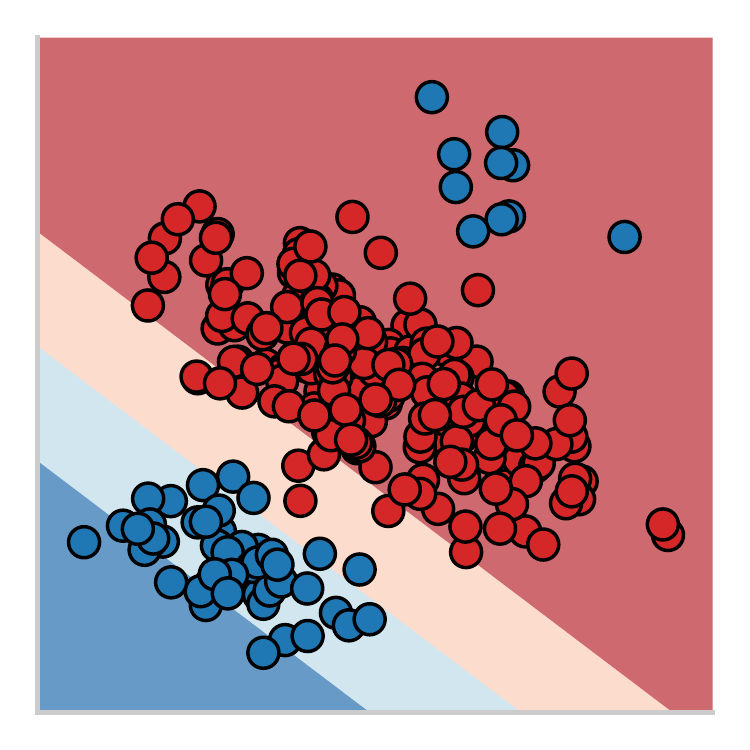}};
    \node[anchor=south] at ($(midleft.north)+(-2.0,0.1)$) {\textbf{Representation comparison}};
    \node[anchor=south] at ($(right.north)+(-2.1,0.1)$) {\textbf{Probe comparison}};

    \node[anchor=south, text width=0.25\textwidth] at ($(left.north)+(1.1,-0.4)$) {\small (a) optimal $R^\ast$};
    \node[anchor=south, text width=0.25\textwidth] at ($(midleft.north)+(1.0,-0.4)$) {\small (b) random $R'$};
    \node[anchor=south, text width=0.25\textwidth] at ($(midright.north)+(1.1,-0.4)$) {\small (c) optimal $P^\ast$};
    \node[anchor=south, text width=0.25\textwidth] at ($(right.north)+(0.8,-0.4)$) {\small (d) insufficient $P'$};
    
    \node at ($(left.south)+(-0,-0.1)$) {
        \begin{tikzpicture}
          \node [draw] {$\scriptstyle \log p(\vproperty | \vtoken, R^\ast, P^\ast)= -53$};
        \end{tikzpicture}
    };
    \node at ($(midleft.south)+(-0,-0.1)$) {
        \begin{tikzpicture}
          \node [draw] {$\scriptstyle \log p(\vproperty | \vtoken, R', P^\ast)= {-516}$};
        \end{tikzpicture}
    };
    \node at ($(midright.south)+(-0,-0.1)$) {
        \begin{tikzpicture}
          \node [draw] {$\scriptstyle \log p(\vproperty | \vtoken, R^\ast, P^\ast)= {-53}$};
        \end{tikzpicture}
    };
    \node at ($(right.south)+(-0,-0.1)$) {
        \begin{tikzpicture}
          \node [draw] {$\scriptstyle \log p(\vproperty | \vtoken, R^\ast, P')= -103$};
        \end{tikzpicture}
    };
    \end{tikzpicture}
    \vspace{-1.5em}
    \caption{
      Comparison of the inductive biases of representation--probe pairs using the \evidence.
      The \evidence below the respective figures indicates that the right probe and representation are selected.
      The probing task is a binary classification of two properties
    (\protect\tikz[baseline=-0.5ex,inner sep=2pt]{\protect\node[C4] {\pgfuseplotmark{*}};}
    vs
    \protect\tikz[baseline=-0.5ex,inner sep=2pt]{\protect\node[C1] {\pgfuseplotmark{*}};}).
    The same colors are used to mark the probe's decision function.
       Representations that naturally separate the properties are preferred over random representations in terms of the \evidence, since they have a better inductive bias.
      \textbf{Left}: we compare an optimal representation that distinguishes both property classes (a) and a random representation (b).
      \textbf{Right}: we compare a neural probe (c) to a linear probe (d) which is too simplistic.
      The \evidence correctly prefers a neural probe since it better explains the data.
    }
    \label{fig:toy}
    \vspace{-1em}
\end{figure*}

\section{Probing as Quantifying Inductive Bias}
\label{sec:method}

At the most fundamental level, the NLP community's interest in pre-trained representations is about reducing the sample complexity of models on downstream tasks.
The community hopes that pre-trained representations
are able to imbue NLP models with enough information about a given language that
models can reach a higher performance with the same or even fewer training data.
And, indeed, over and over again this has been shown to be the case~\citep{ELMO, BERT, T5}.
Another way of phrasing this desire is that the NLP community hopes that pre-trained representations have a suitable inductive bias for downstream tasks.
This paper takes the position that, rather than probing the pre-trained representations for how much linguistic structure they contain---an endeavor that has received much attention~\citep[\emph{inter alia}]{belinkov2017morph,belinkov2018survey,conneau2018sentling,liu2019ling} but is still contentious ~\citep{hewitt2019controlprobe,pimentel2020pareto,pimentel2020infoprobe,voita2020mdlprobing}---we should directly ask how much they improve the inductive bias on tasks of interest.\looseness=-1

We propose to quantify the inductive bias of a model, i.e., a representation--probe pair, using the principle of \defn{Occam's razor}~\citep{rasmussen2001occam}.
Occam's razor states that we should choose the simplest model that sufficiently explains our observations.
One way to operationalize this principle is Bayesian model selection~\citep{rasmussen2001occam, mackay2003information, bishop2006pattern}.
Bayesian model selection relies on the \defn{evidence}, which is a distribution over data sets for a given model---that is, how likely is it that a particular data set could have been generated by that model.
With a probing data set,
the \evidence encompasses Occam's razor because (i) a model that is too simple would assign low probability to the data set (e.g., it is very unlikely that we sample a smooth cubic curve from a linear model), and (ii) an overly complex model would assign low probability because it can model that data set as well as many others (e.g., it is unlikely that we sample a cubic from a deep Transformer). %
In line with Occam's razor, the \evidence is then highest for the simplest model that sufficiently explains the data set (e.g., a cubic model is the best explanation for a data set consisting of a cubic polynomial).

In the following, we outline the probabilistic model for probing and the form of the \evidence.
This enables us to quantify the inductive bias of representations.
Crucially, part of the inference is to select the optimal probe for each representation so as to enable a fair %
comparison between representations.

\subsection{A Probabilistic Model of Probing}

Computation of the \evidence
requires the definition of a probabilistic probing framework.
In this section, we introduce such a framework.
Specifically, we compute the \evidence of representation--probe pairs that constitute models for a fixed task.\footnote{We note that our formulation has a close connection to the MDL formulation of probing~\citep{voita2020mdlprobing}.}

We start by introducing the notation necessary to describe our probabilistic probing framework.
Formally, we denote linguistic sequences by $\token \in \sV^+$, where $\sV$ is a vocabulary.\footnote{$\sV^+$ is the set of all sequences of elements in $\sV$ of length at least $1$.}
For example, $\token$ could be a word in context, a whole sentence, or simply a single token.
We probe for a linguistic property $\property \in \Pi$.
In a probing task, we have a data set of $N$ i.i.d.\ pairs $\{(\token_n, \property_n)\}_{n=1}^N$ of sequences with associated linguistic properties.
We abbreviate all sequences and properties collectively in a data set by $\vtoken$ and $\vproperty$.
Formally, a \defn{representation} $R(\cdot)$ is a (possibly stochastic) function from a sequence to a $D$-dimensional real vector, i.e., $R: \sV^+ \ra \R^D$.
We will use the shorthand $\wordrep = R(\token)$ to represent the vector resulting from the application of the function $R(\cdot)$ to $\token$,
and $\vwordrep$ to abbreviate the representations of all sequences $\vtoken$ in the data set.
Finally, we employ a probe to predict the linguistic property $\property_n$ of a sequence $\token_n$ from its representation $R(\token_n)$, i.e.,
a probabilistic probe $\probe(\cdot)$ maps a vector in $\R^D$ to a distribution over linguistic properties.
In all, this means that the composition $(\probe \circ R)(\token_n)$ yields a distribution over the linguistic property $\property_n$ corresponding to $\token_n$.
As an example, the representation $R(\cdot)$ may be realized by \bert, the probe $\probe(\cdot)$ may be a linear classifier, $\token$ are words in context, and $\property$ are POS tags.\looseness=-1

In our framework, we treat the composition of $\probe(\cdot)$ and $R(\cdot)$ jointly as a single model whose inductive bias we seek to assess.
Formally, we define a model as a \defn{representation--probe pair},
which we denote by a tuple $(R, P) \in \sR \times \sP$, where $R(\cdot) \in \sR$ denotes a representation and $P \in \sP$ is a probe specification.
A \defn{probe specification} characterizes a prior over some family of probes, e.g., a 2-layer neural network probe with tanh activations and a Gaussian prior on the weights.
This is consistent with the probing literature, where probes are often parameterized families of probabilistic models
trained using a regularization scheme that implicitly defines a prior over the parameters.\footnote{For example, L2 regularization can be seen as placing a Gaussian prior over the parameters of a model~\citep[Chapter 7]{murphy2012machine}.}
In such a case, a natural prior has the form $p(\theta \mid \vwordrep, P)$, where $\theta$ are the parameters of the family of models associated with $P$.\footnote{In most applications, we would assume that the prior does not depend on $\vwordrep$, i.e., the prior would simply be $p(\theta \mid P)$. Indeed, we are being more general than what is usually necessary; however, as will soon become clear, allowing for this conditioning will simplify notation.}
Each $P \in \mathcal{P}$ would then specify a prior over probe parameters $\theta$ and thus probe functions $f(\cdot)$.
However, we opt for a slightly different notation. 
Analogous to our notation for $\wordrep$, we define 
$\probe$ for the corresponding vector of probe outputs for an input representation, i.e. $\probe = f(\wordrep)$, and $\vprobe$ as the probe outputs over the entire data set.
Then, we reparameterize the prior $p(\theta \mid \vwordrep, P)$ in terms of the probe outputs $\vprobe$, i.e., $p(\vprobe \mid \vwordrep, P)$.\footnote{This reparameterization may be achieved with a standard application of the change-of-variable formula using the neural network's Jacobian, similar to what is being done in functional variational inference \citep[e.g.,][]{d2021repulsive}.}
Our formulation is therefore general: we can follow previous work on probing and opt for a neural network probe, in which case each $P \in \mathcal{P}$ can specify an architecture and prior over parameters; however, we can also consider priors directly on function outputs, e.g., if we want a Gaussian process probe.

As we mentioned above, we allow for stochastic representations $R(\cdot)$.
We can interpret this as a prior over representation outputs $\wordrep$, which is given by $p(\wordrep \given \token, R)$: it is conditional on the choice of representation and the particular input sequence $\token$ we want a representation for.
Formulating representations as probabilistic allows our framework to be more general, i.e., it can be used to compare stochastic representations~\citep[\emph{inter alia}]{vilnis2015embedding, barkan2017bayesian, xue2021bayesian} to deterministic representations like \bert.
If $R(\cdot)$ prescribes a deterministic representation then the distribution on $\wordrep$ given a sequence $\token$ is given by the Dirac delta function: $p(\wordrep \given \token, R) = \delta(R(\token)-\wordrep)$.

Jointly, the priors over probe and representations outputs specify the prior for a representation--probe pair.
All that remains is specifying the likelihood function; it is defined such that it
factorizes over the data set as
$p(\vproperty \given \vprobe) = \prod_{n=1}^N p(\property_n \given \probe_n)$.
The joint distribution $p(\vproperty, \vprobe, \vwordrep \given \vtoken, R, P)$ of the probabilistic probing model is then given by
\begin{equation}
\label{eq:joint}
    \underbrace{p(\vproperty \given \vprobe)}_{\text{likelihood function}}
    \times \,\,\,
    \underbrace{p(\vprobe \given \vwordrep, P) \,p(\vwordrep \given \vtoken, R)}_{\text{prior}}.
\end{equation}
We obtain the \evidence for our representation--probe tuple by integration:
\begin{align}
\label{eq:marglik}
    p(\vproperty &\given \vtoken, R, P)  \\
    &= \iint p(\vproperty, \vprobe, \vwordrep \given \vtoken, R, P) \, \mathrm{d}\vprobe \, \mathrm{d} \vwordrep.  \nonumber
\end{align}
The \evidence is a distribution over linguistic properties $\vproperty$ given input tokens $\vtoken$ and a particular choice of model, i.e., representation--probe pair $(R, P)$.
A representation--probe pair that could easily generate correct linguistic properties will score a higher \evidence than one that does not generate any linguistically meaningful properties or one that can generate all sorts of data sets.

\subsection{Maximizing the Model Evidence}
To find the best representation--probe pair, we need to find the one maximizing the \evidence in \cref{eq:marglik}:
\begin{equation}
\label{eq:inductive_bias_max}
    (R^\ast, P^\ast) = \argmax_{(R, P) \in \sR \times \sP} p(\vproperty \given \vtoken, R, P).
\end{equation}
The space of representations $\sR$ that we compare when probing is typically quite small and leads to a discrete choice: each $R(\cdot) \in \sR$ simply denotes a distinct choice of representation.
Further, all prior work on probing considers exclusively deterministic representations which, as mentioned above, simplifies the prior over representations to a Dirac delta distribution. 
This means we can rewrite \cref{eq:marglik} as follows
\begin{align}
    \iint &p(\vproperty, \vprobe \given \vwordrep, P) \, \mathrm{d}\vprobe \, \delta(R(\vtoken) - \vwordrep) \, \mathrm{d} \vwordrep \nonumber\\
    = \int &p(\vproperty, \vprobe \given \overline{\vwordrep}_R, P) \, \mathrm{d} \vprobe.
    \label{eq:marglik_rep}
\end{align}
where we use $\overline{\vwordrep}_R = R(\vtoken)$ to emphasize that this is the non-random representation of $\vtoken$ according to $R(\cdot)$.
This characterizes our probing procedure:
we compute this integral independently for each representation $R \in \sR$ and hence the problem in \cref{eq:inductive_bias_max} reduces to selecting, for each representation, the probe specification $P \in \sP$ that maximizes the \evidence.
The \defn{inductive bias} of a representation $R$ is the resulting optimal \evidence across probes: $\max_{P \in \sP} p(\vproperty \given \overline{\vwordrep}_R, P)$. 
This procedure can also be understood as hypothesis testing with a likelihood-ratio test~(see \cref{app:hypothesis_testing}).

While $\sR$ is simply the set of representations that we want to probe, the set $\sP$ that characterizes priors on probes is more complex.
It is typically a combination of discrete and continuous choices:
For example, the number of layers in a neural probe is discrete, but the setting of weight decay is continuous.
Moreover, to ensure that the \evidence is not limited by a restricted choice of probe architectures, the set $\sP$ needs to encompass sufficiently simple and complex probes at the same time.
Hence, we construct our prior on probes by incorporating commonly used probes into it:
we consider linear~\citep{alain2016understanding, adi2016probing, hewitt2019controlprobe, liu2019ling, pimentel2020pareto} and more complex neural probes~\citep{pimentel2020infoprobe, voita2020mdlprobing} paired with weight decay to control complexity~\citep{hewitt2019controlprobe,pimentel2020pareto}.
Probing based on a family of probes instead of a fixed architecture is a key difference to other probing frameworks.
In fact, in our experiments~(\cref{sec:experiments}) we find that different representations perform best with different probe architectures and hyperparameters.
This suggests that limiting probing to a single probe configuration might be misleading. 

In practice, to maximize the \evidence for each representation over $\sP$, we follow the evidence framework by \citet{mackay1995probable, mackay2003information} using the scalable implementation proposed by \citet{immer2021scalable}.
This enables us to quantify the inductive bias of a representation (\cref{eq:marglik_rep}) and maximize it over $P \in \sP$ as required by \cref{eq:inductive_bias_max}, i.e., for each representation we select
$\max_{P \in \sP}p(\vproperty \given \overline{\vwordrep}_R, P)$.
It also allows us to maximize the integral over a set of infinitely many choices of weight decay strength, to further control the complexity of the probes.
As shown in \cref{sec:experiments}, this leads to highly consistent results and alleviates overfitting, which is a problem that even simple linear probes have.

\section{Tackling Probing with Evidence}
\label{sec:wrong-probing}

As outlined in \cref{sec:intro}, current work in probing faces a series of problems.
Here we discuss how these problems are directly addressed
by the \evidence.

\subsection{Problem I (Representation Selection)}
\label{sec:problem-nonsensical}
Clearly, random representations have no suitable inductive bias for linguistic tasks.
Nonsensical results, such as that random representations outperform pre-trained ones~\citep{zhang2018auxiliary, hewitt2019controlprobe, pimentel2020pareto} simply indicate overfitting, which is strictly penalized in our framework.
Compared to pre-trained representations, random representations have low \evidence for linguistic tasks because there is no probe that can reliably predict the properties.
In \cref{fig:toy}a vs.\ \ref{fig:toy}b, we illustrate how a random representation is penalized by the \evidence.
As we will see in \cref{sec:experiments}, our framework consistently assigns lower evidence to the random representations compared to the pre-trained ones.

\subsection{Problem II (Probe Selection)}
\label{sec:problem-linear}
Current probing results are inextricably bound to the choice of probe, yet for probing to provide us with insights about representations, we must break this dependence.
For example, one salient issue in probing is that, while pervasive in the literature, there is a spurious association between linear probes and ease of extraction.
This is illustrated in \cref{fig:toy}, where we can see a linear probe (\cref{fig:toy}d) that offers less ease of extraction than a neural probe (\cref{fig:toy}c), as measured by the \evidence.
This means that could obtain misleading results if we restricted our analysis to linear probes.
Conversely, we will later see that linear probes can be too complex for some probing tasks and overfit, though the \evidence overcomes this problem (\cref{fig:linear-overparam}).
We avoid the problem of selecting a fixed probe by instead choosing a sufficiently large
set $\sP$ of priors of families of probes
and finding the optimal probe specification, within that family, for each representation; as we will see later, the optimal probe varies considerably across tasks and representations.
Instead of heuristic arguments about which probe to choose, the \evidence provides a statistically sound way to select one in line with a likelihood-ratio test~\citep{neyman1933ix}.\footnote{Refer to \cref{app:hypothesis_testing} for more details.}

\subsection{Problem III (Task Selection)}
In our opinion, an important issue with probing is that the research program has unclear goals.
Like much of task-driven NLP, probing is essentially supervised learning with pre-trained representations.
We argue that the goal of quantifying and, in particular, maximizing the inductive bias of representation--probe pairs aligns probing with regular NLP:
In both cases, one searches for an optimal model at the lowest possible complexity---it does not matter whether the task of interest is simple or complex.

\begin{figure*}
    \centering
    \includegraphics[width=\linewidth]{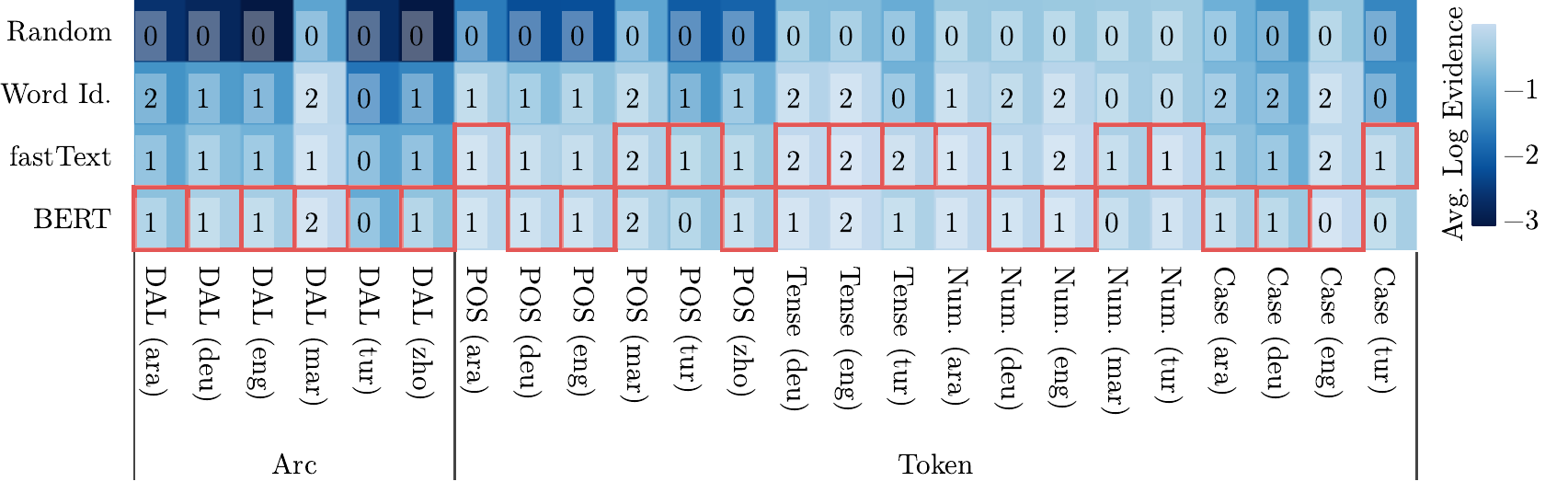}
    \vspace{-2em}
    \caption{Inductive biases of different representations (shown on the rows) for token- and arc-level tasks (shown on the columns), as measured by the log \evidence averaged over the data set.
    The integers inside the cells denote the number of layers of the optimal probe, with 0 denoting a linear probe.
    The representations with the best inductive bias (averaged over 3 runs) for each task are highlighted in red.
    }
    \label{fig:relative-evidence-token-arc}
    \vspace{-1em}
\end{figure*}

\section{Experimental Setup}%
\label{sec:experiments}

We evaluate our framework on a series of token, arc, and sentence tasks.
Our token- and arc-level tasks are multilingual,\footnote{We consider a small but typologically diverse set of languages: English (eng), Arabic (ara), Turkish (tur), Marathi (mar), German (deu), and Chinese (zho).} whereas our sentence tasks only consider English.
We remove any property values that have less than 20 examples in any of the splits.
All our probes are trained using the Adam~\citep{kingma2015adam} optimizer.
For details on hyperparameters, see \cref{app:experimental_details}.

\paragraph{Token-level tasks.}
For our token-level probing tasks, we probe for part-of-speech (POS) tags, tense, number, and case.
We use the setup in \citet{intrinsic}, which consists of mapping the UD v2.5~\citep{ud2.5} treebanks to the UniMorph schema~\citep{kirovUniMorphUniversalMorphology2018} using the converter by~\citet{mccarthyMarryingUniversalDependencies2018}, and extracting examples of tokens tagged for the relevant properties.
Next, we obtain the representations for each of those tokens in their sentential context~\citep{intrinsic}.
Finally, we split the resulting vocabulary using a 65--35 train--test split, such that no word appears in multiple splits.
While the evidence does not require such a split, we use the split to validate results (cf.~\cref{fig:linear-overparam}).

\paragraph{Arc-level tasks.}
For our arc-level tasks, we conduct dependency arc labeling (DAL).
This consists of classifying the label for a dependency relation given only the representations for the head and dependent of that relation.
These are extracted from the UD v2.5 treebanks using the approach in \citet{pimentel2020pareto}. We use the default UD splits.

\paragraph{Sentence-level tasks.}
For our sentence-level tasks, we consider four tasks.
The first is MultiNLI~\citep{multiNLI}, a natural language inference task.
The other three are the BoolQ~\citep{boolq}, Commitment Bank~\citep{commitbank}, and recognizing textual entailment~\citep[RTE;][]{rte1,rte2,rte3,rte5} tasks, which are part of the SuperGLUE benchmark~\citep{superglue}.
If a task requires one or more passages as input, we first obtain a passage-level representations by averaging over all of its tokens.

\paragraph{Representations.}
In our token and arc tasks, we compare four different representations $R \in \sR$:
(i)~m\nobreakdash-\bert~\citep{BERT}, (ii) \fasttext~\citep{fasttext,fasttextMultilingual},
(iii) a random representation (Rand.), which offers no information, drawn i.i.d. from a Gaussian distribution with zero mean and unit variance and the same dimensionality as \bert for each data point, and
(iv) a representation that assigns a unique random vector to every word in our vocabulary, so the only information it provides is the identity of the word (Word Ident.).
The dimensionality of (iii) and (iv) is the same as that of the \bert representation.
For the sentence tasks, we consider (i) Random, (ii) \fasttext, (iii) \bert, (iv) ALBERT~\citep{albert}, (v) RoBERTa~\citep{roberta}, (vi) XLNet~\citep{xlnet}, and (vii) T5~\citep{T5}.
\Cref{app:representations} lists details on the exact models and implementations used.

\paragraph{Probe Family.}
In order to ensure fair comparisons, our framework requires us to define a suitably expressive set of priors $\sP$ over probe families.
In line with most of the probing literature, this includes linear and neural probes with $1$ or $2$ hidden layers, $100$ hidden units, tanh activation, and varying weight decay parameter.

\section{Results}%
\label{sec:results}

\begin{figure}
    \centering
    \includegraphics[width=0.8\linewidth]{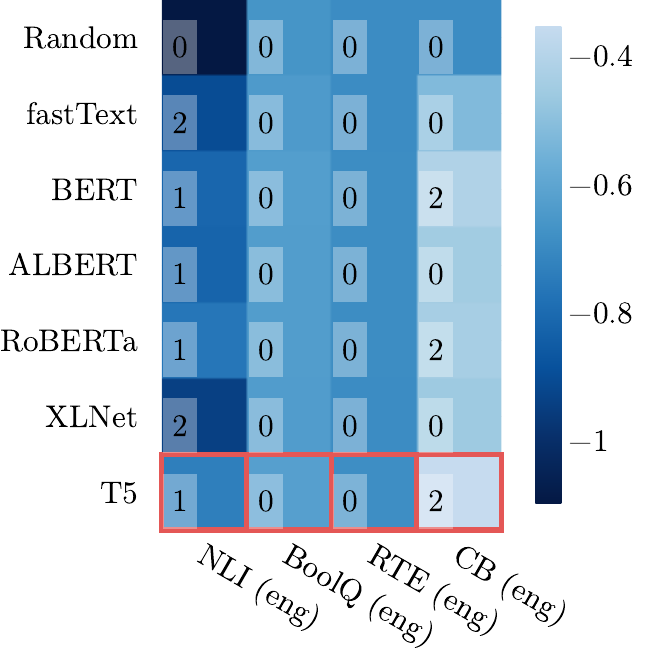}
    \vspace{-1em}
    \caption{Inductive biases of different representations (shown on the rows) for sentence-level tasks (shown on the columns), as measured by the log \evidence averaged over the data set.
    The integers inside the cells denote the number of layers of the optimal probe, with 0 denoting a linear probe.
    The representations with the best inductive bias for each task are highlighted in red.
    }
    \label{fig:relative-evidence-sentence}
    \vspace{-1em}
\end{figure}

We find that our formulation of probing alleviates the problems that we identified in \cref{sec:wrong-probing}.
Firstly, the \evidence suggests that random representations have an unsuitable inductive bias for linguistic tasks, which is in line with hypotheses from previous research~\citep{zhang2018auxiliary,pimentel2020pareto}.
Secondly, the automatic selection of the right probe architecture using the \evidence shows that linear probes are seldom preferred, at least in our token- and arc-level experiments.
That said, we also find evidence that even linear probes can overfit, and that the optimal linear probes may require many of their weights to be regularized to zero.
Clearly, allowing different probe architectures between representations is beneficial for a fair comparison:
simpler representations can profit from a more complex probe and demonstrate a superior inductive bias than more complex representations in some cases.
Specifically, we find that \fasttext demonstrates a better inductive bias than \bert on multiple morphosyntactic tasks, while T5 appears to offer the best inductive bias for all our sentence-level tasks.

\subsection{Representation Comparison}
In the following, we discuss the results presented in \cref{fig:relative-evidence-token-arc} and \cref{fig:relative-evidence-sentence} in detail.

\paragraph{Expected trends.}
Our results depict trends that should be expected from probing.
For example, random representations perform worse than pre-trained representations, especially in tasks with a larger number of classes, such as POS and dependency arc labeling.
Word identity representations are better than random representations, which is to be expected, since
the former are at least able to associate certain types to their most frequent properties, whereas the latter offer no information because they are sampled randomly per token. %
We suspect this is the reason why the optimal probe for random representations is always a linear probe that predicts the majority class.

\paragraph{Token- and arc-level tasks.}
\cref{fig:relative-evidence-token-arc} contains the results of our token- and arc-level tagging tasks.
We find that \fasttext offers a better inductive bias for tense, while \bert is superior for case across all languages with the exception of Turkish (tur).
In fact, we find that \fasttext evinces a better inductive bias for all Turkish token-level tasks.
We believe that this is due to the agglutinative nature of Turkish, which means that \fasttext's bag-of-subword-units mechanism provides a useful inductive bias.
For dependency arc labeling (DAL), we find that \bert has a uniformly better inductive bias.
Interestingly, other than for random representations, the optimal probe usually has a non-linearity, which refutes the idea that linear probes should be blindly picked for their simplicity.
In all, our token- and arc-level results
suggest that \bert is not a panacea, and motivate further research into multilingual studies of the morphosyntactic properties that \bert exposes well.

\paragraph{Sentence-level task.}
\cref{fig:relative-evidence-sentence} suggests that \tfive~\citep{T5} has a better inductive bias than the other representations we consider on sentence-level tasks.
That said, we find that the difference in \evidence between the different representations is generally quite small for \boolq, \rte, and \commit.
Indeed, despite these being highly complex tasks, a linear probe is uniformly preferred for \boolq and \rte.
This may be an indication that the sentence-level representation mechanism we chose, i.e., averaging over the representations for the tokens in a sentence, is particularly ineffective for these two tasks.
Indeed, we see that for both tasks, the evidence for the representations is not much higher than the evidence for the random representation, which may indicate that the optimal probes are largely ignoring the representations and just learning a majority-class baseline, which is achieved at the smallest complexity using a linear probe.

\begin{figure}
    \centering
    \includegraphics[width=0.99\columnwidth]{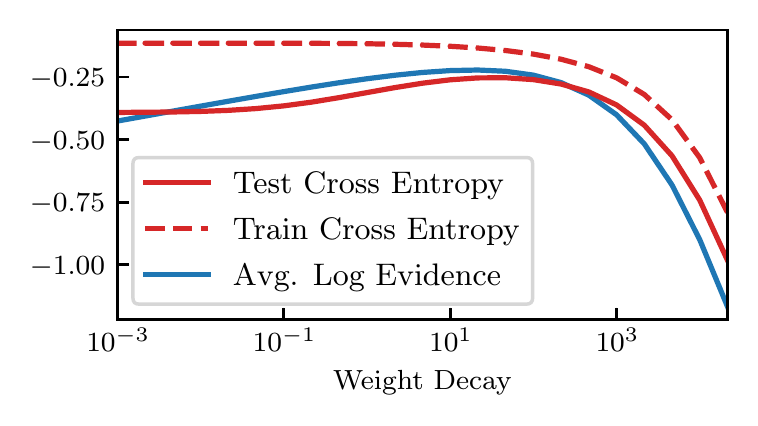}
    \vspace{-2em}
    \caption{Average $\log$ \evidence and cross entropy (\emph{higher is better}) versus weight decay of a linear probe on POS German tasks using \bert. 
    Even a simple linear probe can overfit and yield misleading results if not regularized properly as we can see from the generalization gap between training
    and test 
    cross-entropies.
    }
    \label{fig:linear-overparam}
    \vspace{-1em}
\end{figure}

\subsection{Controlling Probe Complexity}
\label{sec:results-overfitting}
\cref{fig:linear-overparam} shows linear probes on two tasks and how the \evidence and cross-entropy change as a function of their weight decay.
The graph shows that insufficient regularization leads to poor generalization using \bert, apparent from the gap between training and test loss that grows larger when  weak regularization is applied.
This means that insufficiently regularizing linear probes---and hence allowing them to fully use their parameters---reduces their \evidence.

\begin{figure}
    \centering
    \includegraphics[width=0.99\columnwidth]{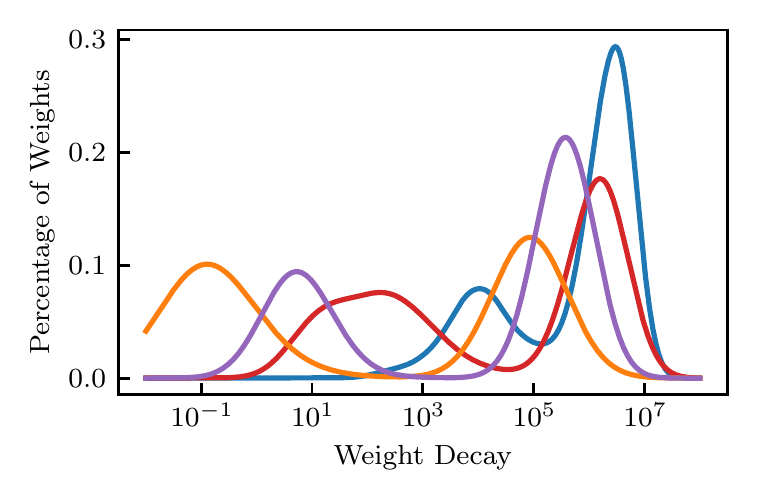}
    \vspace{-2em}
    \caption{Distribution of the prior precisions learned by linear English POS probes for \textcolor{C1}{\textbf{Random}}, \textcolor{C4}{\textbf{Word Identity}},
    \textcolor{C2}{\textbf{\fasttext}}, and
    \textcolor{C5}{\textbf{\bert}} representations. High regularization causes weights to be effectively zeroed out.}
    \label{fig:bimodal}
    \vspace{-1em}
\end{figure}

This observation, alongside former results, led us to conjecture that optimal probes may actually be restricted linear models, i.e., linear probes where most parameters are disabled.
Our implementation is easily able to account for this hypothesis: by expanding $\sP$ so that each parameter gets associated a different regularization strength, we can \emph{automatically} identify which parameters are needed and force others towards zero.
\cref{fig:bimodal} illustrates the resulting distribution of per-parameter regularization strengths in the optimal probe for English POS, when $\sP$ is defined to be the set of linear probes with per-parameter regularization;
interestingly, the distribution is bimodal, such that every representation has a set of parameters that is zeroed out (rightmost mode).
The random representation is regularized more than pre-trained ones, because it can only learn a majority baseline.
Note that in practice, we can do this for probes with multiple layers too, so that the optimal probe we find may be simultaneously deep and sparse.

\section{Related Work}
\label{sec:related}

Probing aims to provide insights into what linguistic information is encoded in pre-trained representations.
Since the introduction of probing for sentence representations~\citep{adi2016probing, conneau2018sentling}, probing has also been applied to representations of words and tokens~\citep{belinkov2018survey, liu2019ling, voita2020mdlprobing, pimentel2020infoprobe}.
Nonetheless, comparison of representations, the choice of probe, and even probing tasks have been under scrutiny recently~\citep{belinkov2018survey, liu2019ling, hewitt2019controlprobe, pimentel2020infoprobe}.

\paragraph{Measuring representation quality.}
Prior work has mostly used probe accuracy as a measure of the quality of a representation.
However, if not properly cross-validated, this can lead to nonsensical results which suggest that random representations are as good as learned ones~\citep{zhang2018auxiliary, hewitt2019controlprobe}.
To alleviate this problem, control tasks~\citep{hewitt2019controlprobe}, fewer data~\citep{zhang2018auxiliary}, or simplistic probes~\citep{liu2019ling} have been used.
Using the \evidence can be seen as extensive cross-validation~\citep{fong2020marginal} and is therefore better suited for comparing representations.

In recent work, \citet{lovering2021predicting} argue that the ease of extraction of relevant features can be seen as an inductive bias.
Specifically, they present experiments on artificial and naturalistic tasks that suggest that the amount of fine-tuning data required to make models rely on relevant features as opposed to spurious correlates of the output is connected to the relative ease of extraction between the spurious and relevant features.
In comparison, our method can be seen as integrating over the entire space of features that a representation offers, and as such makes no assumptions about how a task should be solved, i.e., whether certain features are spurious or not for the task at hand.

\paragraph{Simple or complex probes?}
The choice of probe architecture is still a point of contention in the literature.
Initially probes were typically linear models~\citep{alain2016understanding, adi2016probing, liu2019ling} because complex probes could memorize and overfit~\citep{zhang2018auxiliary, hewitt2019controlprobe}.
However, restricting ourselves to linear probes only allows us to ask whether a particular task has a linear decision boundary, which tells us little about the information encoded in representations.
Therefore, neural probes have recently been used as well~\citep{pimentel2020infoprobe, voita2020mdlprobing}.
In particular, this has spawned a line of work on automatically trading off probe performance and complexity.
For example, \citet{hewitt2019controlprobe} propose control tasks that mitigate overfitting and find that weight decay helps generalization in line with our observations in \cref{sec:results-overfitting}.
\citet{voita2020mdlprobing} use the minimum description length (MDL) principle which is equivalent to the \evidence in the case of a probabilistic model~\citep{mackay2003information}.
Both of these frameworks focus on the comparison and selection of probes which we argue is distinct from the problem of comparing representations.
Thus in our framework, two representations do not need to be compared using the same probe but on the basis of the optimal probe for the representation, which appears to be useful~(\cref{sec:results}).
In this sense, our work is most similar to \citet{pimentel2020pareto}, where representations, as opposed to probes, are compared by considering the Pareto hypervolume.
That said, their approach is dependent on the choice of a complexity metric, whereas ours is not.

\paragraph{Linear probes can overfit.}
Our results indicate that, for some tasks, even linear probes may be overparameterized. 
One possible reason for this is that the optimal probes for these tasks ignore portions of the representation.
If true, this would suggest that our framework may be useful for neuron-level probing~\citep{dalvi2019neuron, durrani2020neurons, intrinsic, antverg2022neurons}, whose goal is to identify subsets of neurons in a representation that are informative about a property of interest.

\section{Conclusion}

Previous approaches to linguistic probing are plagued by several key problems, namely the issues of nonsensical results, probe selection, and ill-defined goals.
To overcome these issues, we have proposed a novel probing framework, which focuses on the inductive bias that pre-trained representations offer for different linguistic tasks.
We have shown that the Bayesian \evidence, a natural measure for inductive bias, can be used in the context of probing.
We have found that our framework empirically does not suffer from the aforementioned problems.
We are hopeful that under this new paradigm, future work in probing will be more principled, comparable, and useful to the NLP community at large.

\section*{Ethics Statement}
The authors foresee no ethical concerns with the work presented in this paper.

\section*{Acknowledgements}
The authors thank Tiago Pimentel, Karolina Sta\'{n}czak, and members of the McGill NLP group for discussions and providing feedback in various stages of this project, and the anonymous reviewers for their valuable feedback.
A.\ I.\ acknowledges funding by the Max Planck ETH Center for Learning Systems (CLS).
L.\ T.\ H.\ acknowledges funding from the Michael Athans Fellowship fund.
V.\ F.\ acknowledges funding through a PhD fellowship from the Swiss Data Science Center.
R.\ C.\ acknowledges support from the Swiss National Science Foundation (SNSF) as part of the ``The Forgotten Role of Inductive Bias in Interpretability'' project.

\bibliography{references}
\bibliographystyle{acl_natbib}

\clearpage
\appendix

\onecolumn

\section{Probing as Hypothesis Testing}
\label{app:hypothesis_testing}
The result of maximizing over probe $P$ and representation $R$ in \cref{eq:inductive_bias_max} can also be understood as selecting the highest-scoring hypothesis in a likelihood-based hypothesis test~\citep{neyman1933ix}.
To see this, we treat the pair $(R, P)$ as one variable $M$.
Then, we compare two representation--probe pairs, $M$ and $M'$, which we both believe to be equally probable, in the light of data by their likelihood ratio
\begin{align}
    \frac{p(M \given \vproperty, \vtau)}{p(M' \given \vproperty, \vtau)}
    &= \frac{p(\vproperty \given \vtau, M) \, p(M) \, p(\vproperty \given \vtau)}{p(\vproperty \given \vtau, M') \, p(M') \, p(\vproperty \given \vtau)} \nonumber \\
    &= \frac{p(\vproperty \given \vtau, M)}{p(\vproperty \given \vtau, M')}.
\end{align}
The simplification is due to the equal probability of $M$ and $M'$, i.e., $p(M) = p(M')$.
If the resulting ratio is larger than one, we decide in favor of $M$, otherwise in favor of $M'$.
Thus, for a larger set $\sP$ of probes and $\sR$ of representations, it is sufficient to identify the highest-scoring pair in terms of the evidence $p(\vproperty \given \vtau, M)$, which is exactly the objective in \cref{eq:inductive_bias_max}.

\section{Experimental Details}
\label{app:experimental_details}
All our probes are trained using the Adam~\citep{kingma2015adam} optimizer with hyperparameters $\beta_1 = 0.9, \beta_2 = 0.999$, learning rate $0.1$, batch size $512$, and for $500$ epochs.
For each discrete architecture (linear, MLP-1, MLP-2), we run the \evidence framework as suggested by \citet{immer2021scalable} with the following parameters: frequency $F=1$, $K=100$ number of steps every epoch, learning rate $\gamma=0.1$.
The evidence framework provides a Laplace approximation to the evidence in \cref{eq:marglik_rep}.
We implement our probing method using \texttt{laplace-torch}~\citep{daxberger2021laplace} and use a Kronecker-factored Hessian approximation~\citep{martens2015optimizing} using the eigendecomposition for the Laplace approximation~\citep{immer2021improving}.
We use an individual weight decay parameter per parameter group of the probes, i.e., each set of weights and biases are regularized independently per layer.
Only for \cref{fig:bimodal}, we use a weight decay strength individually per parameter of the linear probe which effectively turns off individual parameters by increasing their weight decay.
This is also known as automatic relevance determination~\citep{mackay1995probable}.
In this case, we use a diagonal Hessian, and thus posterior, approximation.

\section{Representations}
\label{app:representations}
\cref{tab:representations} shows the representations we used.
For all transformer models, we use the HuggingFace transformers library~\citep{wolfHuggingFaceTransformersStateoftheart2020}.
Note that for \fasttext we use the multilingual vectors which are language-dependent, and the official \fasttext library.\footnote{\url{https://pypi.org/project/fasttext/}}

\begin{table}[h!]
\small
    \centering
    \begin{tabular}{@{}ll@{}}
    \toprule
    Representation & Model name                   \\ \midrule
    m-BERT         & \texttt{bert-base-multilingual-cased} \\
    BERT           & \texttt{bert-base-uncased}            \\
    fastText       & Language-specific, see \href{https://fasttext.cc/docs/en/crawl-vectors.html}{here}.                       \\
    T5             & \texttt{t5-base}                      \\
    RoBERTa        & \texttt{roberta-base}                 \\
    XLNet          & \texttt{xlnet-base-cased}             \\
    ALBERT         & \texttt{albert-base-v2}               \\ \bottomrule
    \end{tabular}
    \caption{Representations used. All representation except \fasttext use the HuggingFace implementations~\citep{wolfHuggingFaceTransformersStateoftheart2020}.}
    \label{tab:representations}
\end{table}

\end{document}